\documentclass[11pt]{article}
\usepackage{eacl2017}
\usepackage{times}
\usepackage{url}
\usepackage{latexsym}
\usepackage{graphicx}

\eaclfinalcopy 

\title{Inherent Biases of Recurrent Neural Networks for Phonological Assimilation and Dissimilation}

\author{Amanda Doucette \\
  University of Massachusetts Amherst \\
  {\tt amandakdoucette@gmail.com}\\}

\date{}

\begin{document}
\maketitle
\begin{abstract}

A recurrent neural network model of phonological pattern learning is proposed. The model is a relatively simple neural network with one recurrent layer, and displays biases in learning that mimic observed biases in human learning. Single-feature patterns are learned faster than two-feature patterns, and vowel or consonant-only patterns are learned faster than patterns involving vowels and consonants, mimicking the results of laboratory learning experiments. In non-recurrent models, capturing these biases requires the use of alpha features or some other representation of repeated features, but with a recurrent neural network, these elaborations are not necessary.

\end{abstract}

\section{Introduction}

Models of phonological pattern learning typically require large numbers of constraints or rules on where features can occur, and the presence of alpha features or some other representation of repeated features to allow certain patterns to be learned more quickly~\cite{hayes2008,moreton2015}. In human learning experiments, certain phonological patterns are learned more easily, particularly those involving multiple occurences of the same feature, such as a voicing agreement pattern. 

In order to capture this bias towards single-feature patterns, many models have some representation of repeated features. Alpha features are one example of this (see McCarthy~\shortcite{mccarthy1988} for other approaches, such as feature geometry). Alpha features allow a model to learn a harmony pattern with only one predicate - that two features must be the same, having the value $\alpha$. Without alpha features, the model must learn two predicates - that the two features must either both have the value $+$ or the value $-$. Therefore, there cannot be a bias towards single-feature patterns, because two-feature patterns also require learning two predicates~\cite{moreton2012}.

In addition to alpha features, many phonological learning models have to test or search over a large number of possible rules or constraints to learn a pattern. In models that use conjunctions of features as constraints~\cite{hayes2008,moreton2015}, if there are $N$ features in the model, each with three possible values ($+, -, \pm$), there are $3^N$ possible conjunctions of these features. With even a small number of features, the number of conjunctive constraints becomes very large.

Moreton, Pater, and Pertsova~\shortcite{moreton2015} describe a cue-based learning model that uses these conjunctive constraints. Their model is a maximum entropy model trained by gradient descent on negative log-likelihood, and is related to the single-layer perceptron. It successfully models the biases found in human phonological learning experiments, but still requires listing all possible constraint conjuncions in the input. In unpublished work, I have found that it is also possible to model these biases without constraint conjunctions using a feed-forward neural network with a hidden layer. See Alderete and Tupper~\shortcite{alderete} for an overview of other connectionist approaches to phonology.

Hare~\shortcite{hare1990} uses a recurrent neural network to model Hungarian vowel harmony without phonological rules or constraints. In Hare's model, sequences of individual features describing vowels were the only inputs to the network. Some features in the input sequence could be left unspecified, and after training, fully specified feature sequences are output. While the model was only trained on sequences of vowels, not entire words, Hare showed that recurrent neural networks were capable of modeling vowel harmony patterns using only individual features as input.

Rodd~\shortcite{rodd1997} also uses recurrent neural networks to model Turkish vowel harmony. Individual phonemes rather than features were used as input to the networks, and the task was to predict the following phoneme. Rodd showed that the hidden units in small recurrent networks were able to represent distinctions between vowels and consonants, differences in sonority, and differences between front and back vowels. Although humans most likely do not perform the task of predicting the next phoneme in a word, Rodd showed that simple recurrent network could learn phonological regularities through differences in the distribution of phonemes.

Recurrent neural networks are capable of learning more than just vowel harmony patterns and feature representations, though. This paper describes a simple recurrent neural network model of phonological pattern learning that is biased towards learning single-feature patterns and patterns over only consonants or vowels without using alpha features, separate representations of consonants and vowels, or conjunctive constraints.

\section{Model}

The model used in these simulations is a simple recurrent neural network model. The "words" that make up the patterns are the inputs in the first layer of the model. At each time step, the four features representing one phoneme are input to the network. The second layer is a hidden recurrent layer with ten neurons. The third layer is a log softmax layer with two output neurons. After the entire sequence is input to the model, the outputs at the final timestep will represent the log probabilities of the input belonging to each of the two classes, which will be referred to as IN the pattern or OUT of the pattern. The probability of a pattern being IN or OUT is the probability of it being allowed in the language.

The model was trained using gradient descent on negative log-likelihood with a learning rate of 0.01. Weights were adjusted after each word in the training data, rather than in batches. In each epoch of training, the order of presentation of the training data was randomly permuted. The output of the network was considered to be correct when the log probability of the intended class was greater than the other class. This criterion for correctness was used because when the model is trained on a subset of the full pattern, it prevents overfitting to that subset. After each epoch of training, all training examples were checked for correctness. If the correct class was predicted for every training example, training was stopped.

The number of neurons in the recurrent layer is not important to the model. Ten neurons were chosen because with fewer neurons, the patterns tested could not be fully learned, and with more, the patterns were learned after only a few training epochs. More complex patterns or patterns requiring more features will likely require a larger number of neurons in this layer.

\section{Patterns}

The patterns used in testing the model used four phonological features - two consonant features, and two vowel features. Each feature has a value of +1 or -1. For consonants, the vowel features have a value of 0, and for vowels, the consonants have a value of 0. The consonant features used in these patterns are voicing ($+/-$voi) and place($+/-$cor), and the vowel features are height ($+/-$hi) and backness ($+/-$back). This feature set corresponds to the consonants [d, t, k, g] and the vowels [i, u, \ae, a]. All "words" in the patterns have the form $C_1V_1C_2V_2$, where $C$ and $V$ range over the four consonants and vowels described by the four features, so there are 256 total. For each pattern, the words are divided into the two classes, IN and OUT, each with 128 examples.

Six patterns, dividing the 256 words based on their features, were created as simplified versions of real phonological patterns. The six feature combinations that are IN for each pattern are described in Table 1. The 128 words that do not fit these feature descriptions are in the OUT class of the pattern.

\begin{table}[h]
\begin{center}
\begin{tabular}{|l|l|}
\hline \bf Pattern & \bf Features in Pattern \\ \hline
\bf 1 & C1+voi and V1+back\\
& C1-voi and V1-back\\
\hline
\bf 2 & C1+voi and V2+back\\
& C1-voi and V2-back\\
\hline
\bf 3 & C1+voi and C2+voi\\
& C1-voi and C2-voi\\
\hline
\bf 4 & C1+voi and C2-voi\\
& C1-voi and C2+voi\\
\hline
\bf 5 & C1+voi and C1+cor\\
& C1-voi and C1-cor\\
\hline
\bf 6 & C1+voi and C2+cor\\
& C1-voi and C2-cor\\
\hline
\end{tabular}
\end{center}
\caption{\label{pattern-table} Feature descriptions of patterns. }
\end{table}

In pattern 1, there is a feature dependency between an adjacent consonant and vowel. In pattern 2, this dependency is between a non-adjacent consonant and vowel. Pattern 3 is a voice assimilation pattern where the two consonants must agree in voicing. In pattern 4, the consonants must disagree in voicing. In pattern 5, the two features relevant to the pattern are on the same consonant, and in pattern 6, they are on two separate consonants.

Moreton~\shortcite{moreton2012} claims that there is an advantage for learning intra-dimensional patterns over inter-dimensional patterns that requires alpha features to be captured by a model. The same advantage was also shown by Moreton, Pater, and Pertsova~\shortcite{moreton2015} and Saffran and Thiessen~\shortcite{saffran2003}. In the six patterns described here, patterns 3 and 4 are intra-dimentional, single-feature patterns, and the rest are inter-dimensional, two-feature patterns. Therefore, patterns 3 and 4 should be learned faster than patterns 1, 2, and 6. 

Pattern 5 is also an inter-dimensional pattern, but the two features are on the same segment rather than different segments like the rest of the patterns. In the experiments of Moreton, Pater, and Pertsova~\shortcite{moreton2015}, patterns involving features on the same segment were easier to learn than patterns involving two segments. Because of this, pattern 5 should be learned faster than pattern 6.

Moreton~\shortcite{moreton2012} also showed in an experiment that a pattern involving an adjacent consonant and vowel was learned no faster than a pattern involving a non-adjacent consonant and vowel. Therefore, there should be no difference in the amount of time to learn patterns 1 and 2.

Results from several studies also show that there is no difference in difficulty of learning harmony and disharmony patterns~\cite{moreton2012,pycha2003,skoruppa2011}. Of these six patterns, pattern 3 is a harmony pattern, and pattern 4 is a disharmony pattern, so there should be no difference in the time to learn these patterns.

\section{Results: Training on full patterns}

For each pattern, the model was trained 3000 times with random initial weights on all 256 examples. For each training run, the number of epochs taken to learn the pattern according to the criterion in section 2 was recorded. Table 2 shows averages over the 3000 runs for each pattern. The model was capable of learning the patterns in all but 25 training runs, which were stopped after 400 epochs and excluded from these results. This was done because weights in these training runs were likely stuck in local minima, and the model was incapable of learning the pattern.

\begin{table}[h]
\begin{center}
\begin{tabular}{|l|l|l|l|}
\hline \bf Pattern & \bf Mean & \bf St. Err. & \bf St. Dev. \\ \hline
1 & 22.03 & 0.22 & 11.78\\ \hline
2 & 22.74 & 0.26 & 14.49\\ \hline
3 & 19.91 & 0.29 & 15.88\\ \hline
4 & 20.20 & 0.29 & 16.08\\ \hline
5 & 19.65 & 0.15 & 8.42\\ \hline
6 & 22.46 & 0.23 & 12.79\\
 \hline
\end{tabular}
\end{center}
\caption{\label{allResults-table} Number of epochs to learn patterns with full training set. }
\end{table}

In a two sample t-test, there is no evidence that there is a significant difference between the number of epochs taken to learn patterns 3 and 4, the harmony and disharmony patterns (p \textgreater 0.01). Patterns 3 and 4, the one-feature patterns, were learned significantly faster than patterns 1, 2, and 6, the two-feature patterns involving two segments in two sample t-tests (p \textless 0.01). However, there was no difference between the one-feature patterns 3 and 4, and pattern 5, which involved two features on the same segment (p \textgreater 0.01). Pattern 5 was also learned faster than pattern 6, which involved two features on different consonants (p \textless 0.01). There was also no difference between pattern 2, which involved a non-adjacent consonant and vowel, and pattern 1, which involved an adjacent consonant and vowel (p \textgreater 0.01). Individual comparisons and exact p-values are shown in Table 3.

\begin{table}[h]
\begin{center}
\begin{tabular}{|l|l|}
\hline \bf Comparison & \bf p-value \\ \hline
Pattern 1, Pattern 2 & $0.03934$\\ \hline
Pattern 5, Pattern 6 & $2.2\times 10^{-16}$\\ \hline
Pattern 1, Pattern 3 & $4.091\times 10^{-9}$\\ \hline
Pattern 2, Pattern 3 & $6.505\times 10^{-13}$\\ \hline
Pattern 4, Pattern 3 & $0.4843$\\ \hline
Pattern 5, Pattern 3 & $0.4361$\\ \hline
Pattern 6, Pattern 3 & $8.932\times 10^{-12}$\\ \hline
Pattern 1, Pattern 4 & $4.583\times 10^{-7}$\\ \hline
Pattern 2, Pattern 4 & $1.414\times 10^{-10}$\\ \hline
Pattern 5, Pattern 4 & $0.1006$\\ \hline
Pattern 6, Pattern 4 & $1.909\times 10^{-9}$\\
 \hline
\end{tabular}
\end{center}
\caption{\label{pvalue-table} Training time comparisons for pairs of patterns and p-values. }
\end{table}

\section{Results: Training on subset of patterns}

The model was also trained on randomly chosen subsets of the training data for each pattern. This was done because when learning phonological patterns, people do not have access to every possible example of the pattern, and every possible example of something that does not conform to the pattern. Rather, people are only exposed to correct examples of their language as they learn.

To test the model under these conditions, 32 examples of each pattern were randomly chosen from the 128 available. If the model were trained on only positive examples, it would predict everything to be included in the pattern, so some negative training examples are needed. As negative training examples, 32 were chosen from the remaining 224, after removing the 32 already selected positive examples. Therefore, the negative examples are not true negative examples, they are just randomly chosen examples from the full dataset.

Although this is an unusual method of training a neural network model, it was done to test if the model is capable of generalizing to unseen examples of the pattern. There is no direct way of using unsupervised learning with a recurrent neural network, so using randomly selected negative examples was used as an analogue to it. In training, the model gets positive examples of the pattern, but does not get true negative examples.

The model was trained 3000 times on randomly chosen subsets of each pattern. Training was stopped when the model correctly classified the 64 examples it was trained on. After each training run, the proportion of the full dataset that was classified correctly by the model was recorded. Table 3 shows averages over the 3000 runs for the number of training epochs taken and proportion correct in the full dataset for each pattern. 41 training runs were excluded because the pattern was not learned in 400 epochs.

\begin{table}[h]
\begin{center}
\begin{tabular}{|l|l|l|l|l|l|}
\hline \bf Patt. & \bf Mean & \bf Mean & \bf Std. Err. & \bf Std. Dev. \\
& \bf Corr. & & & \\ \hline
1 & 0.73 & 125.6 & 0.67 & 36.6\\ \hline
2 & 0.64 & 131.1 & 0.69 & 37.6\\ \hline
3 & 0.72 & 133.8 & 0.77 & 42.3\\ \hline
4 & 0.72 & 130.8 & 0.72 & 39.3\\ \hline
5 & 0.76 & 125.7 & 0.72 & 39.6\\ \hline
6 & 0.69 & 125.0 & 0.66 & 36.0\\
 \hline
\end{tabular}
\end{center}
\caption{\label{allResults-table} Number of epochs to learn patterns with partial training set. }
\end{table}

When trained on a subset of the pattern, the mean number of epochs taken to learn each pattern did not differ in the way they did when trained on the entire pattern. The mean number of epochs taken is similar for each pattern, with a much larger standard deviation than when the model was trained on the full pattern. This is likely because the subsets were randomly chosen. Sometimes the negative examples in the subset would conflict with the pattern represented by the positive examples, resulting in much longer training times.

While the training times do not follow the predicted pattern, these results show that the model is still able to generalize the pattern to examples it is not trained on, as shown in the mean proportion correct column in Table 3. The model was trained on 25\% of the pattern data, only 12.5\% of which were positive examples of the pattern, but can correctly classify 64\% to 76\% of the examples.

\section{Discussion and conclusion}

Without a representation of repeated features, and using only single features as input, this recurrent neural network is able to model results from human phonological learning experiments. Although non-recurrent neural network models such as the single-layer perceptron require a representation of repeated features to allow single-feature patterns to be learned more easily~\cite{moreton2012}, the addition of a recurrent layer seems to have the same effect. In a recurrent neural network, input is processed one segment at a time, rather than simultaneously. There is no clear way to represent sequential input in non-recurrent models. Although it is difficult to interpret connection weights in a recurrent neural network, it is possible that the sequential input of the network somehow biases it towards single-feature patterns.

A non-recurrent, multilayer perceptron may be able to learn these patterns because they are all the same length, but it will not be able to capture the bias towards patterns with repeated features. By concatenating the features of all four segments into one input to a multilayer perceptron, there is no connection between segments with repeated features; a pattern involving two instances of a consonant feature will be connected to the hidden layer in the same way a pattern with a consonant and a vowel will be. In a recurrent network, repeated activation of the same input feature will increase activation in a hidden unit more than activation of different features in the input will. For this reason, a non-recurrent network cannot model biases towards single-feature patterns.

In addition to the bias towards single-feature patterns, the recurrent neural network can learn patterns over only consonants or only vowels more quickly than patterns involving a consonant and a vowel. Some models accomplish this with an additional representation of only the vowels in a word, or only the consonants in a vowel or consonant 'tier'~\cite{hayes2008}. A vowel tier allows for a bias towards vowel-only patterns because only the vowels in a word are considered, making finding the pattern much faster. 

This recurrent neural network model does not have any representation of a consonant or vowel tier, but it is still able to learn consonant-only patterns faster. There are separate features used for describing vowels and consonants, but they are all used as input to the network at the same time. This bias is possibly related to the single-feature bias. Vowels and consonants never have the same features, so patterns with only consonants or vowels are learned faster because there is some overlap in the features used to describe them. It is possible that adding a separate representation of vowels and consonants could make learning these patterns faster, but it does not seem necessary. However, this could be accomplished by having multiple instances of the model where only consonants, only vowels, or the entire word is input. The outputs of the recurrent layers of these separate instances would be connected to a single non-recurrent layer which would combine the predictions of the three tiers.

Although the patterns used to test the model can be described with only one or two features, it should also be capable of learning more complex patterns. More neurons in the input and recurrent layers will allow more input features to be used, and more complex patterns to be represented.

The model also requires supervised training. To learn a pattern, both positive and negative examples are necessary for the model, but humans are capable of learning these patterns through unsupervised learning~\cite{moreton2015}. Unsupervised learning can be approximated by using randomly chosen examples from the entire dataset as negative examples, but it is still not true unsupervised learning. If there is a way to better approximate unsupervised learning with this model, it would better fit the human learning experiments.

In conclusion, a simple recurrent neural network was able to model human phonological pattern learning without alpha features or any representation of rules or constraints. The model uses only individual phonological features as input, and has no separate representations of vowels or consonants. Single-feature patterns are learned more easily than two-feature patterns, and vowel or consonant-only patterns are learned more easily than patterns involving vowels and consonants. Modeling these biases using a recurrent neural network is possible without any representation of repeated features that is necessary in non-recurrent models.

\section*{Acknowledgments}

Thanks to Joe Pater and Brendan O'Connor for their guidance and helpful discussion while I worked on this research, as well as to the four anonymous reviewers for their comments.

\bibliography{paper_final}
\bibliographystyle{eacl2017}

\end{document}